\RequirePackage[loading]{tracefnt}
\documentclass[conference]{IEEEtran}
\usepackage{fancyhdr}

\IEEEoverridecommandlockouts

\usepackage{cite}
\usepackage{amsmath,amssymb,amsfonts}
\usepackage{algorithmic}
\usepackage{enumitem}
\usepackage{graphicx}
\usepackage{textcomp}
\usepackage{xcolor}
\usepackage{soul}
\usepackage{url}
\usepackage{subfigure}
\usepackage{subcaption}
\usepackage{adjustbox}
\usepackage{footnote}
\usepackage{multirow}
\definecolor{airforceblue}{rgb}{0.36, 0.54, 0.66}
\definecolor{green}{rgb}{0.490,0.757,0.655}
\definecolor{blue}{rgb}{0.569,0.624,0.780}
\definecolor{lightorange}{rgb}{0.965,0.843,0.729}
\definecolor{orange}{rgb}{0.929,0.576,0.420}

\makeatletter
\newcommand*{\rom}[1]{\expandafter\@slowromancap\romannumeral #1@}
\makeatother
\def\BibTeX{{\rm B\kern-.05em{\sc i\kern-.025em b}\kern-.08em
    T\kern-.1667em\lower.7ex\hbox{E}\kern-.125emX}}
\begin{document}

\title{\textsc{Long Exposure}: Accelerating Parameter-Efficient Fine-Tuning for LLMs under Shadowy Sparsity}
\author{
\IEEEauthorblockN{
Tuowei Wang\IEEEauthorrefmark{1}\IEEEauthorrefmark{2}\IEEEauthorrefmark{3}\thanks{\IEEEauthorrefmark{3} Work done during an internship at Microsoft Research.}\IEEEauthorrefmark{4}\thanks{\IEEEauthorrefmark{4} Tuowei Wang and Kun Li have equal contributions to this article.},
Kun Li\IEEEauthorrefmark{1}\IEEEauthorrefmark{4}\IEEEauthorrefmark{5}\thanks{\IEEEauthorrefmark{5} Corresponding authors (kunli@microsoft.com; renju@tsinghua.edu.cn).},
Zixu Hao\IEEEauthorrefmark{1}\IEEEauthorrefmark{2},
Donglin Bai\IEEEauthorrefmark{1}, \\
Ju Ren\IEEEauthorrefmark{2}\IEEEauthorrefmark{5},
Yaoxue Zhang\IEEEauthorrefmark{2},
Ting Cao\IEEEauthorrefmark{1},
Mao Yang\IEEEauthorrefmark{1}
}
\IEEEauthorblockA{\IEEEauthorrefmark{1} Microsoft Research\\ Beijing, China}
\IEEEauthorblockA{\IEEEauthorrefmark{2} Tsinghua University\\ Beijing, China}
}

\maketitle

\thispagestyle{fancy}
\lhead{}
\rhead{}
\chead{}
\lfoot{\footnotesize{
SC24, November 17-22, 2024, Atlanta, Georgia, USA
\newline 979-8-3503-5291-7/24/\$31.00 \copyright 2024 IEEE}} \rfoot{}
\cfoot{}
\renewcommand{\headrulewidth}{0pt} \renewcommand{\footrulewidth}{0pt}

\begin{abstract}
The adaptation of pre-trained large language models (LLMs) to diverse downstream tasks via fine-tuning is critical for numerous applications. However, the inefficiency of parameter-efficient fine-tuning (PEFT) techniques presents significant challenges in terms of time investments and operational costs. In this paper, we first introduce a nuanced form of sparsity, termed \textit{Shadowy Sparsity}, which is distinctive in fine-tuning and has not been adequately addressed for acceleration. Under Shadowy Sparsity, we propose \textsc{Long Exposure}\footnote{Long Exposure is available at \url{https://github.com/HPHEX/LongExposure}.}, an efficient system to accelerate PEFT for LLMs. \textsc{Long Exposure} comprises three key components:  \textit{Shadowy-sparsity Exposer} employs a prolonged sensing range to capture more sparsity details under shadowy sparsity; \textit{Sequence-oriented Predictor} provides efficient yet accurate predictions to handle large sequence inputs and constantly-evolving parameters; and \textit{Dynamic-aware Operator} facilitates more structured computational patterns and coalesced memory accesses, addressing dynamic sparse operations. Extensive evaluations show that \textsc{Long Exposure} outperforms state-of-the-arts with up to a $2.49\times$ speedup in end-to-end fine-tuning, offering promising advancements in accelerating PEFT for LLMs.
\end{abstract}


\section{Introduction}
In natural language processing, the adaptation of pre-trained large language models (LLMs)~\cite{gpt3,llama,llama2,opt,palm} to diverse downstream tasks constitutes a fundamental aspect of many applications. This adaptation process, commonly known as \textit{fine-tuning}, involves the comprehensive update of all parameters within the pre-trained model akin to training from scratch.

For the potential hundreds of thousands of downstream applications that rely on LLMs, the efficiency of fine-tuning directly affects their operational costs and time investments. Given that pre-trained LLMs need periodic updates, typically every few months, to integrate the latest knowledge, there is a pressing demand for accelerating the LLM fine-tuning process.

\begin{table}[h!]\small
    \caption{OPT-1.3B fine-tuning time breakdown. (ms/batch)}
    \label{tab:breakdown}
    \begin{adjustbox}{width=0.50\textwidth,center}
    \begin{tabular}{lllll}
    \hline
       Phase                         & Forward       & Backward      & Optim. Step  & Total \\ \hline
       Full Param.                   & 112.8(27.7\%) & 223.7(54.9\%) & 70.6(17.3\%) & 407.2 \\
       LoRA~\cite{lora}              & 135.3(40.4\%) & 196.3(58.7\%) &  2.0(0.6\%)  & 334.6 \\
       Adapter~\cite{adapter-1}      & 123.6(42.2\%) & 168.4(57.5\%) &  0.7(0.3\%)  & 292.9 \\
       Bitfit~\cite{bitfit}          & 117.6(40.5\%) & 172.4(59.4\%) &  0.2(0.07\%) & 290.3 \\
       P-Tuning~\cite{prefix-tuning} & 137.5(40.1\%) & 193.9(56.6\%) & 11.1(3.2\%)  & 342.6 \\
    \hline
    \end{tabular}
    \end{adjustbox}
\end{table}

The major reason hindering the fine-tuning efficiency is the retention of the same number of parameters in the new model as in the original one. Efforts have been made to address this concern by introducing \textit{parameter-efficient fine-tuning} (PEFT) techniques~\cite{peft-survey}, which only selects or injects a minimal number of parameters for adaption to new tasks. One prominent approach in the domain of PEFT is low-rank adaption (LoRA)~\cite{lora}. It freezes pre-trained model weights and injects smaller, trainable low-rank matrices into each transformer block. Compared to full fine-tuning, LoRA decreases the number of trainable parameters to less than 0.01\%.



This substantial reduction in the number of trainable parameters mitigates the need for maintaining and updating the optimizer states for most parameters. However, PEFT techniques fall short of achieving an expected decrease in wall-clock time. As detailed in Table~\ref{tab:breakdown}, even with minimal parameters being trainable, techniques like LoRA only experience an 18\% reduction in wall-clock time. While PEFT techniques notably cut down the optimization step's wall-clock time, they leave the duration of the forward and backward passes either unchanged or slightly increased. This is because, despite most pre-trained parameters being frozen, computing gradients for trainable parameters still requires complete forward and backward passes through the backbone model. Consequently, the forward and backward passes have emerged as the computational bottlenecks impeding further acceleration.

In this paper, we propose \textsc{Long Exposure}\footnote{Similar to employing a slow shutter speed in photography to capture more light for producing clearer images, this paper adopts a series of granular techniques to expose and leverage more sparsity for accelerating fine-tuning.}, an efficient system to accelerate parameter-efficient fine-tuning for LLMs. The design of \textsc{Long Exposure} is grounded in a crucial observation that PEFT and inference in LLMs exhibit high similarities in their computation patterns. In PEFT techniques, a majority of model parameters remain frozen, similar to the scenario in model inference where parameters also stay unaltered. Previous studies~\cite{dynamic-sparsity-survey,spatten,energon,reformer,longformer,big-bird,informer,local-attn,dynamic-sparse-attention,deja-vu,powerinfer,llm-in-a-flash,pit} have evidenced that LLMs typically exhibit considerable sparsity, with a great number of activations can be excluded from computation to expedite inference in wall-clock time while preserving quality.
Guided by this observation, the key insight of \textsc{Long Exposure} is inspired: given the striking similarities in computation patterns between PEFT and inference, \textit{why not build a bridge to PEFT acceleration by capturing intrinsic sparsity like inference?}

\begin{figure}
    \centering
    \includegraphics[width=0.50\textwidth]{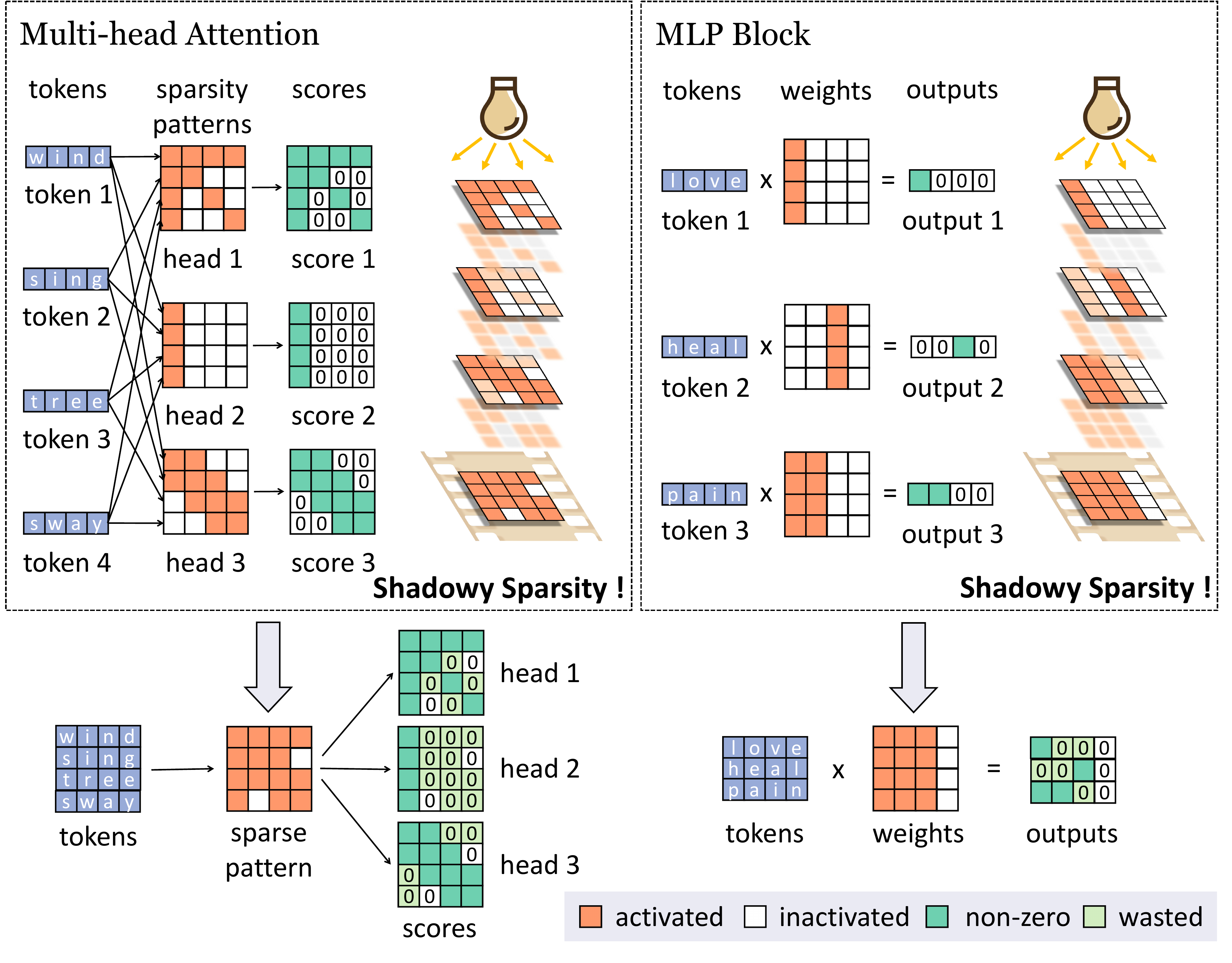}
    \caption{Shadowy sparsity in LLM fine-tuning. Transformer-based models are generally composed of two primary components: multi-head attention and MLP block. In fine-tuning, the sparse patterns of various tokens within the input sequence exhibit a logical AND relationship, which restricts the sparsity degree and leads to computational waste in both components.}
    \label{fig:shadowy-sparsity}
\end{figure}


However, this is not a low-hanging fruit, as the sparsity inherent in fine-tuning introduces distinct characteristics that diverge significantly from those encountered during inference. In inference, the model typically processes one token at a time, where the sparse pattern is easily discernible for each token. In contrast, fine-tuning involves feeding the model with a sequence of tokens, where the sparsity patterns heavily overlap across different tokens, as depicted in Figure~\ref{fig:shadowy-sparsity}. We coin this intricate sparsity observed in fine-tuning as \textit{Shadowy Sparsity}. To accelerate PEFT for LLMs under this shadowy sparsity, several key technical challenges must be tackled carefully.

Firstly, \textit{how to capture more sparse patterns under shadowy sparsity}. As illustrated in Figure~\ref{fig:shadowy-sparsity}, the sparse pattern of the input token sequence emerges from the logical AND combination of the sparse patterns of individual tokens in the sequence. This means that the dense units for a specific token might coincide with the sparse units for another token. The resulting shadowy sparsity exhibits a limited level of overall sparsity, despite the high sparsity degree of each token, leading to potential computational waste.

 

Secondly, upon capturing sparsity, the subsequent critical challenge lies in \textit{how to predict efficient yet accurate sparse patterns} to minimize associated computational expenses before incurring actual costs. Due to the exact sparse patterns typically varying with different inputs, a commonly employed method in inference is to utilize neural networks for predicting sparse patterns at runtime~\cite{dynamic-sparse-attention,deja-vu,powerinfer,llm-in-a-flash}. However, directly flattening the token sequence in fine-tuning as network inputs could lead to an excessively large network, which is both memory-intensive and time-consuming. Moreover, the continuously-evolving trainable parameters during fine-tuning also add complexity to ensuring the correctness of the prediction.


Thirdly, \textit{how to achieve effective performance improvements based on well-predicted sparsity}. The irregular computation patterns and scattered memory accesses associated with sparsity make it challenging to attain comparable performance improvements to the theoretical computation reductions. Moreover, these well-predicted sparsity patterns exhibit highly dynamic characteristics that vary with different inputs at runtime. This renders many existing tools ineffective in capturing and handling such dynamic variations.

\textsc{Long Exposure} employs a suite of techniques to address these challenges. The concept of `\textsc{Long Exposure}' emphasizes that rather than simply harnessing the limited sparsity remaining in shadowy sparsity, we take a longer view which captures more intricate details of individual sparse pattern before they fade into shadow. The core of \textsc{Long Exposure} is the \textit{Shadowy-sparsity Exposer}, a technique designed for exposing the latent sparsity hidden in shadowy sparsity. In multi-head attention, we introduce specific sparse patterns tailored to each attention head, avoiding the computational redundancy or oversight that can arise from employing a uniform mask. In MLP block, we take the importance of each activated neuron into consideration. By identifying and filtering out neurons whose activation can be safely disregarded, we transform shadowy sparsity into structured block-wise sparsity.

\textsc{Long Exposure} utilizes \textit{Sequence-oriented Predictors} to address the conflicts between long sequence inputs and the associated neural network size. This technique is grounded in a two-stage design strategy: Initially, the predictor processes each token individually; then these predictions are subsequently consolidated. Moreover, to minimize the disruption caused by updating trainable parameters, we introduce specific training optimizations to bolster the predictor's robustness.

\textsc{Long Exposure} develops a collection of \textit{Dynamic-aware Operators} to facilitate practical acceleration on hardware systems, covering all the sparse operations involved in multi-head attention and MLP block. Different from most existing tools, these operators avoid additional data conversion overhead, making them well-suited for dynamic scenarios. In addition, we design a two-stage algorithm for multi-head attention that adeptly balances precomputation with dynamic sparse patterns.

We evaluate \textsc{Long Exposure} across various PEFT methods and on two different GPU platforms. The results show that our system achieves up to $2.49\times$ speedup and $2.77\times$ memory savings in end-to-end fine-tuning compared with the state-of-the-art fine-tuning system, maintaining model accuracy.

In summary, our contributions are as follows:
\begin{itemize}
    \item We are the first to identify and leverage the intrinsic sparsity within LLM fine-tuning, namely shadowy sparsity, to accelerate the PEFT process for LLMs.
    \item We introduce three key components that capture, predict, and exploit sparsity patterns, respectively. This approach provides a coherent strategy for optimizing both the multi-head attention and the MLP block within LLMs.
    \item We implement these techniques as an end-to-end fine-tuning system that is compatible with a variety of PEFT techniques. Our system achieves up to $2.49\times$ speedups and $2.77\times$ memory savings compared to the state-of-arts.
\end{itemize}


\section{Background and Motivation}
\subsection{Parameter-efficient Fine-tuning (PEFT)}
Adapting a large pre-trained language model for various downstream applications typically involves full fine-tuning, where all parameters of the pre-trained model are updated. However, as models grow in size, full fine-tuning has evolved from being inconvenient to almost impractical. To reduce the costs associated with full fine-tuning, various PEFT methods have emerged in recent years. The central concept behind these methods is to avoid updates of the full set of parameters without performance degradation. 

A promising direction within PEFT involves freezing the pre-trained model's parameters while introducing a small number of new trainable parameters. One such method~\cite{adapter-1, adapter-2, adapter-3} involves the use of adapters, which are additional layers inserted between the model's existing layers. Specially, Low-Rank Adaptation (LoRA)~\cite{lora} injects trainable low-rank matrices into each layer, reflecting the insight that the updates to model weights actually operate within a low intrinsic dimension. Other methods~\cite{prompt-tuning, prefix-tuning} use prompts, adding trainable parameters to the model's input to leverage the pre-trained model's existing knowledge for new tasks. Instead of introducing new parameters, some methods~\cite{diff-pruning, bitfit} selectively update a small portion of the pre-trained model's parameters, such as only the bias terms~\cite{bitfit}. However, owing to the inherent computational flow of backpropagation, the significant reduction of trainable parameters provided by PEFT primarily benefits the optimizer step, leaving the forward and backward phases as the new bottlenecks.

\subsection{Sparsity in LLMs}

A significant number of activations within both two primary components of transformer-based LLMs: the multi-head attention and the MLP block, are found zero or nearly zero~\cite{pit,sparse-mlp}. These negligible activations can be disregarded with no or little impact, leading to sparsity in model's computational demands. In multi-head attention, sparsity typically emerges from the limited interactions among different tokens. Models can selectively mask out irrelevant tokens, leading to less computation without accuracy degradation. In MLP block, sparsity is primarily attributed to the properties of the ReLU activation function, which is increasingly utilized by many LLMs~\cite{opt,relu-llama,relu-in-llm}. This function sets all negative activation values to zero, allowing them to be excluded from computations.

This inherent sparsity inspires us to leverage it for expediting the parameter-efficient fine-tuning of LLMs. Unlike the original model training, the entire parameters of pre-trained model are frozen during PEFT. This allows for accurate predictions of sparse patterns in both multi-head attention and MLP block, surpassing the limitations of pre-defined sparse attention masks~\cite{reformer,longformer,big-bird,informer,local-attn}, which may not be suitable for all inputs. Although some research~\cite{deja-vu,powerinfer,llm-in-a-flash} has employed similar techniques to accelerate LLM inference, the unique characteristics of sparsity during fine-tuning present substantially different challenges. To our knowledge, no research has utilized this sparsity to accelerate the fine-tuning process.

\subsection{Analysis: PEFT Computational Cost Breakdown}
Fine-tuning a pre-trained model with trainable parameters consists of three phases: (1) the forward phase calculates the loss for current data batch; (2) the backward phase calculates the gradients of the trainable parameters; (3) the optimizer step updates the trainable parameters using these gradients.

\begin{figure}
    \centering
    \includegraphics[width=0.50\textwidth]{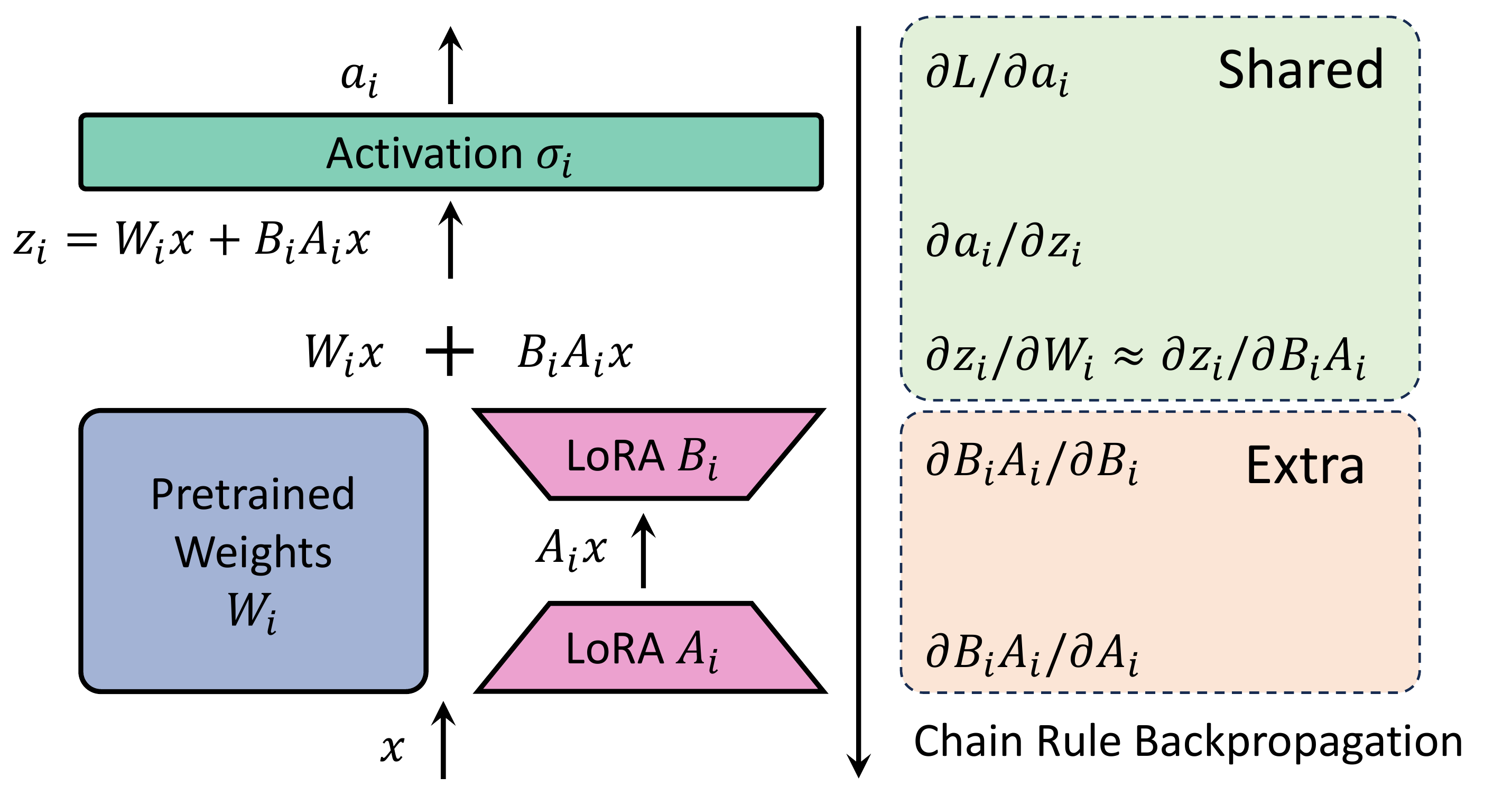}
    \caption{An illustrated example of LoRA for showcasing the computational flow (forward and backward) in PEFT.}
    \label{fig:analysis}
\end{figure}

Consider the use of LoRA to fine-tune an MLP block with alternating linear and activation layers, as shown in Figure~\ref{fig:analysis}. The $i$th linear layer consists of weight $W_{i}$, LoRA matrices $A_i$ and $B_i$, and the $i$th activation layer is $\sigma_{i}$. For $i$th layer, we denote the output of $i$th linear layer as $z_{i}$, and the output following activation as $a_{i}$.

During the forward phase, the computational costs of PEFT are either unchanged or slightly increased. Taking our LoRA example, the output of $i$th linear layer is computed as follows:
\begin{equation*}
    z_{i} = W_{i}x + B_{i}A_{i}x
\end{equation*}
\begin{equation*}
    a_{i} = \sigma_{i}(z_{i})
\end{equation*}
Compared to the original process, the injected LoRA matrices $A_i$ and $B_i$ slightly increase the computational costs.

During the backward phase, the situation is more complex. In backpropagation with loss $L$ of full fine-tuning, the LoRA matrices $A_i$ and $B_{i}$ are absent, and the gradient for the trainable weight $W_{i}$ is:
\begin{equation*}
    \frac{\partial L}{\partial W_{i}} = \frac{\partial L}{\partial a_{i}}\frac{\partial a_{i}}{\partial z_{i}}\frac{\partial z_{i}}{\partial W_{i}}
\end{equation*}

In contrast, when using LoRA, the gradients relative to the trainable parameters $A_{i}$ and $B_{i}$ are:
\begin{equation*}
    \frac{\partial L}{\partial A_{i}} = \frac{\partial L}{\partial a_{i}}\frac{\partial a_{i}}{\partial z_{i}}\frac{\partial z_{i}}{\partial B_{i}A_{i}}\frac{\partial B_{i}A_{i}}{\partial A_{i}}
\end{equation*}
\begin{equation*}
    \frac{\partial L}{\partial B_{i}} = \frac{\partial L}{\partial a_{i}}\frac{\partial a_{i}}{\partial z_{i}}\frac{\partial z_{i}}{\partial B_{i}A_{i}}\frac{\partial B_{i}A_{i}}{\partial B_{i}}
\end{equation*}

Though using LoRA can skip the calculation of $\frac{\partial z_{i}}{\partial W_{i}}$, it introduces the calculations of $\frac{\partial z_{i}}{\partial B_{i}A_{i}}\frac{\partial B_{i}A_{i}}{\partial A_{i}}$ and $\frac{\partial z_{i}}{\partial B_{i}A_{i}}\frac{\partial B_{i}A_{i}}{\partial B_{i}}$ instead. Given that the computational costs of $\frac{\partial z_{i}}{\partial W_{i}}$ are comparable to $\frac{\partial z_{i}}{\partial B_{i}A_{i}}$, LoRA leads to additional computational steps involving $\frac{\partial B_{i}A_{i}}{\partial A_{i}}$ and $\frac{\partial B_{i}A_{i}}{\partial B_{i}}$. Apart from these differences, the remaining computational costs are essentially equivalent as a result of the chain rule, which both require traversing the layers of the extensive pre-trained model.

In the optimizer step, the computational costs of PEFT are reduced due to fewer trainable parameters. However, the extent of this saving can vary depending on the choice of optimizer and typically accounts for a small fraction of the overall. In summary, PEFT methods do not reduce the computational cost significantly compared to full fine-tuning.
\subsection{Opportunity: Accelerate PEFT with LLM Sparsity}
\label{sec:opportunity}
The application of PEFT techniques has greatly cut down the cost of the optimizer update phase, turning the forward and backward phases into new bottlenecks. Unlike standard fine-tuning or training, where model parameters are continuously updated through iterations, PEFT techniques maintain most of the parameters frozen. This closely mirrors the computational pattern observed during inference. Given that sparsity is commonly employed to diminish computational expenses during inference, there is a compelling opportunity to apply sparsity to streamline the fine-tuning as well. However, fine-tuning inherently involves both the forward and backward phases, while inference is limited to the forward phase. To successfully adopt sparsity in fine-tuning, it is necessary to analyze how sparsity impacts the backward phase.

Considering the same example shown in Figure~\ref{fig:analysis}, the output of the $i$th linear layer can be expressed as follows, with $W_{i}$ denoting the $i$th row of $W$:
\begin{equation*}
\begin{aligned}
    & z = Wx + BAx = (W + \Delta W)x \\
    & = \begin{pmatrix} (W_{1} + \Delta W_{1})x, \hdots, (W_{i} + \Delta W_{i})x, \hdots, (W_{d} + \Delta W_{d})x \end{pmatrix}^{T}
\end{aligned}
\end{equation*}

To incorporate sparsity, we suppose $z_{i} = (W_{i} + \Delta W_{i})x < 0$ and set the activation function as $\text{ReLU}$. Consequently, $W_{i}$ is inactivated and the activation can be expressed as:
\begin{equation*}
    a = \text{ReLU}(z) = \begin{pmatrix} (W_{1} + \Delta W_{1})x, \hdots, 0, \hdots, (W_{d} + \Delta W_{d})x \end{pmatrix}^{T}
\end{equation*}

In backpropagation, the gradient relative to $z$ can be expressed as:
\begin{equation*}
\begin{aligned}
    \frac{\partial L}{\partial z} 
    & = \text{ReLU}^{'} \odot \frac{\partial L}{\partial a} \\
    & = \begin{pmatrix} 1, \hdots, 0, \hdots, 1 \end{pmatrix}^{T} \odot \begin{pmatrix} \frac{\partial L}{\partial a_{1}}, \hdots, \frac{\partial L}{\partial a_{i}}, \hdots, \frac{\partial L}{\partial a_{d}} \end{pmatrix}^{T} \\
    & = \begin{pmatrix} \frac{\partial L}{\partial a_{1}}, \hdots, 0, \hdots, \frac{\partial L}{\partial a_{d}} \end{pmatrix}^{T}
\end{aligned}
\end{equation*}

To calculate the gradients relative to the trainable parameters $A_{i}$ and $B_{i}$ for updating LoRA matrices, we first calculate the gradient relative to their product:
\begin{equation*}
\begin{aligned}
    \frac{\partial L}{\partial B_{i}A_{i}} 
    & = \frac{\partial L}{\partial z}x^{T} \\
    & = \begin{pmatrix} \frac{\partial L}{\partial z_{1}}, \hdots, 0, \hdots, \frac{\partial L}{\partial z_{d}} \end{pmatrix}^{T} \begin{pmatrix} x_{1}, \hdots, x_{i}, \hdots, x_{d} \end{pmatrix} \\
    & = \begin{pmatrix}
        \frac{\partial L}{\partial z_{1}}x_{1} & \hdots & \frac{\partial L}{\partial z_{1}}x_{i} & \hdots & \frac{\partial L}{\partial z_{1}}x_{d} \\
        \vdots & & \vdots & & \vdots \\
        0 & \hdots & 0 & \hdots & 0  \\
        \vdots & & \vdots & & \vdots \\
        \frac{\partial L}{\partial z_{d}}x_{1} & \hdots & \frac{\partial L}{\partial z_{d}}x_{i} & \hdots & \frac{\partial L}{\partial z_{d}}x_{d} \\
        \end{pmatrix} \\
\end{aligned}
\end{equation*}
This gradient does not involve $\frac{\partial L}{\partial z_{i}}$, implying that no corresponding $W_{i}$ is involved either. It leads to an important conclusion: \textit{if certain model parameters remain inactive during the forward phase, they are effectively excluded from the gradient computation in the backward phase}. This conclusion highlights the potential for leveraging sparsity to decrease the computational demands of PEFT, analogous to the efficiencies observed during inference.


\section{Overview}
\begin{figure}
    \centering
    \includegraphics[width=0.50\textwidth]{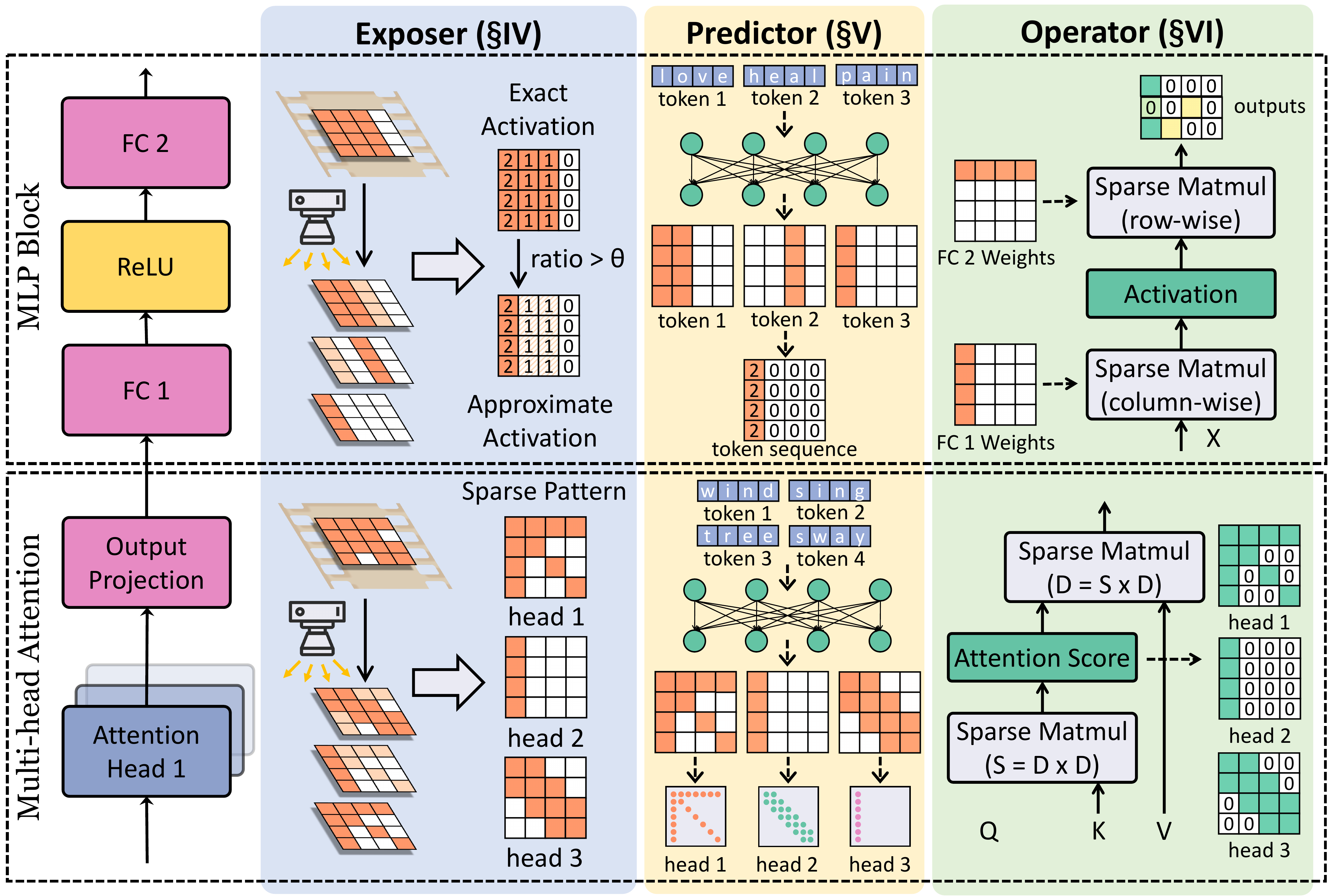}
    \caption{\textsc{Long Exposure} overview}
    \label{fig:overview}
\end{figure}

We propose \textsc{Long Exposure}, an efficient fine-tuning system with LLM instinctive sparsity. Beyond prior works that concentrate exclusively on either parameter-efficiency during fine-tuning or computation-efficiency during inference, \textsc{Long Exposure} is efficient on aspects of both parameter and computation. Figure~\ref{fig:overview} presents an overview of our system.

\noindent\textbf{Shadowy-sparsity Exposer (Section \ref{sec:exposer}).} Drawing from our experimental findings, we present several key observations of LLM sparsity in fine-tuning. Different from inference, the sparse patterns in fine-tuning are heavily overlapping across different tokens. We refer to this as \textit{Shadowy Sparsity}. To expose the latent sparse patterns inherent in shadowy sparsity, we adopt a more granular approach. In multi-head attention, we consider the unique features of each head and employ a head-specific sparse mask. In MLP block, we factor in the importance of activated neurons and introduce a neuron-filter to transform shadowy sparsity into block-wise sparsity.

\noindent\textbf{Sequence-oriented Predictor (Section \ref{sec:predictor}).} \textsc{Long Exposure} employs neural-network-based predictors to determine the desired sparse patterns at runtime. In multi-head attention, the goal is to predict the sparse mask for each head. In MLP block, the prediction targets the neurons that are activated in the model weights. To control the size of predictors, we initially process each token in the sequence individually and then combine the outputs. Data argument techniques and tailored loss metrics are introduced to increase predictor accuracy in the presence of trainable parameters during fine-tuning.

\noindent\textbf{Dynamic-aware Operator (Section \ref{sec:operator}).} \textsc{Long Exposure} integrates a suite of dynamic-aware operators that make efficient use of predicted sparse patterns. In multi-head attention, each head is associated with a distinct sparse pattern, which necessitates two sparse matrix multiplications. We design a two-stage algorithm that shifts the bulk of the data format conversion overhead in matrix multiplication to pre-runtime, while still meeting the dynamic nature of sparse patterns. In MLP block, model parameters are activated either by row or by column, i.e. a neuron. We optimize the standard block-wise matrix multiplication by taking into account the neuron-wise sparse pattern. Additionally, we optimize the data layout to enhance memory coalescing.

\section{Shadowy-Sparsity Exposer}
\label{sec:exposer}
\subsection{Observation: Shadowy Sparsity}
Many studies~\cite{dynamic-sparsity-survey,dynamic-sparse-attention,spatten,energon,sparse-mlp} have highlighted the inherent sparsity found in LLMs. Moreover, the sparsity within LLM is dynamic, indicating that the sparse patterns change with different inputs. Most of the existing works~\cite{deja-vu,powerinfer,llm-in-a-flash} concentrate on model inference, where the model input during decoding is a single token (sequence length is 1). However, we observe a distinct characteristic of sparsity during LLM fine-tuning, where the input is a sequence of tokens.

\begin{figure}
    \centering
    \includegraphics[width=0.50\textwidth]{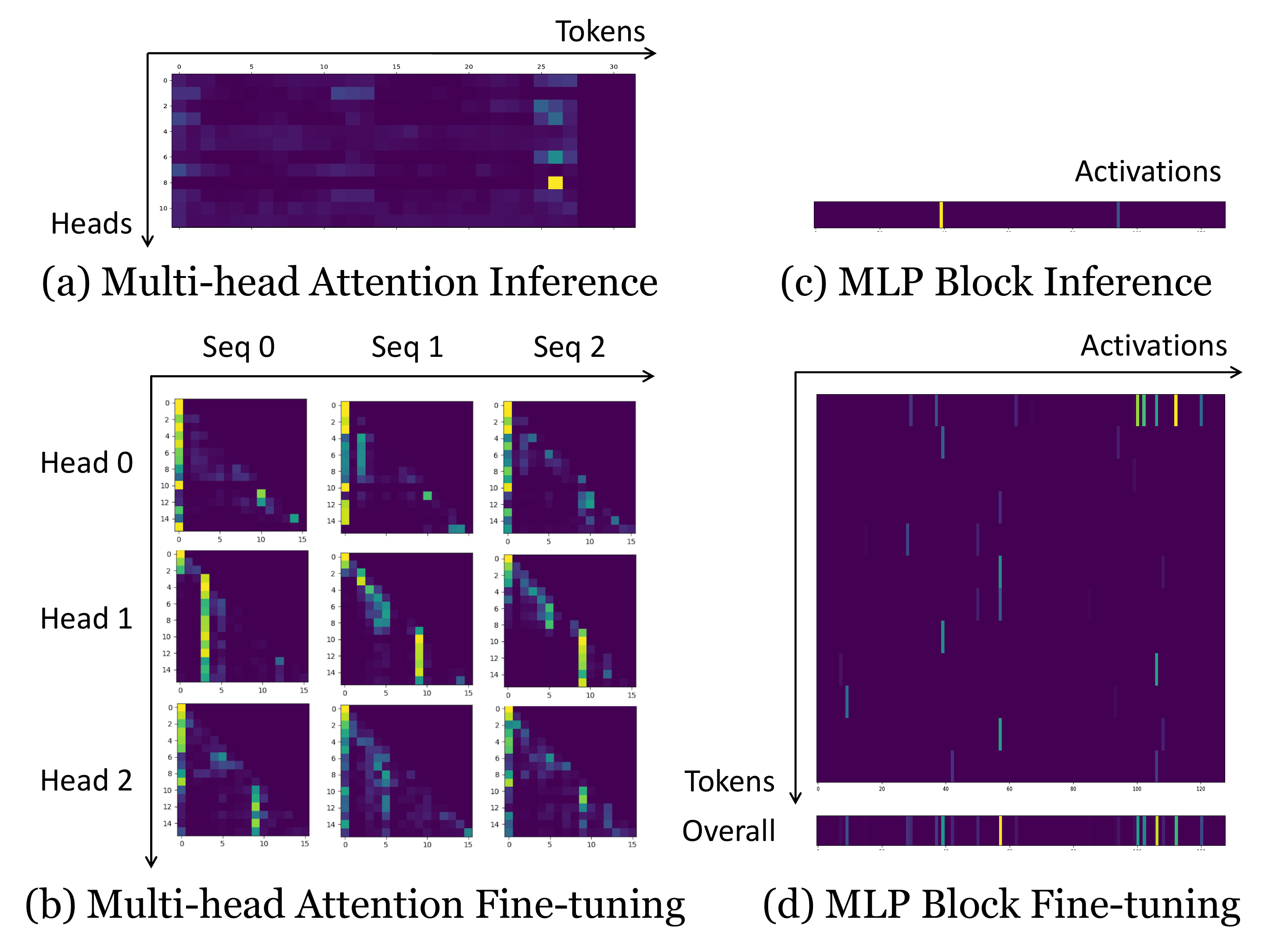}
    \caption{Visualization of attention scores in multi-head attention and activations in MLP block. Due to varying input lengths, these metrics manifest as vectors during inference and as matrices during fine-tuning. Brighter colors denote high values.}
    \label{fig:observation}
\end{figure}

\noindent\textbf{Multi-head Attention.} Figure~\ref{fig:observation}(a) shows the attention scores across different heads for an example token. The sparsity in multi-head attention during inference is characterized as certain heads giving heavy attention scores while others are rather uniform. This pattern offers an opportunity to selectively focus on those `heavy hitter' heads while disregarding the rest. However, during fine-tuning, the attention scores become a matrix that describes the relationships among all tokens in the sequence, as shown in Figure~\ref{fig:observation}(b). It becomes challenging to prune any particular head since each head might be activated by a certain token in the sequence. Several studies~\cite{reformer,longformer,big-bird,informer,local-attn} suggest the use of sparse masks, designed to retain all critical attention scores for saving attention computation. However, existing sparse masks are typically pre-defined and uniformly applied to all heads, which fails to efficiently capture the various sparsity patterns in both the head and input dimensions. 

\noindent\textbf{MLP Block.} A similar phenomenon is observed in MLP block. Figure~\ref{fig:observation}(c) shows the activations after applying ReLU for a single token, which exhibits considerable sparsity. However, with a sequence of tokens as input, the sparsity of overall activations, which is measured by the reduction along the sequence length dimension, reduces greatly as shown in Figure~\ref{fig:observation}(d). Besides, these scattered activations render the residual sparsity more unstructured, posing a challenge for leveraging it for practical acceleration.

We refer to this intricate sparsity observed in fine-tuning as \textit{Shadowy Sparsity}, akin to each token casting a partial shadow, and when these shadows overlap, no light shines through at all. The presence of shadowy sparsity during fine-tuning makes it challenging to capture efficient sparsity patterns. On one hand, shadowy sparsity reduces the degree of sparsity, rendering it inadequate for achieving practical speedups. On the other hand, the remaining sparsity is typically highly unstructured, which is unaligned with the hardware characteristics. 

\subsection{Design: Long Exposure}
To capture more sparse patterns under shadowy sparsity, our primary design lies in focusing on the intricate details of individual sparse pattern that constitutes the shadowy sparsity. We term this as \textit{Long Exposure}, which entails employing a long-duration sensing range to capture more sparsity details.

\noindent\textbf{Multi-head Attention.} We employ a binary mask to capture the inter-token sparsity within a sequence. Each element of the mask corresponds to a block of attention scores, with 0 indicating non-computation and 1 indicating computation. Different from previous studies, our approach operates at the level of individual attention heads, instead of the entire multi-head attention. Concretely, for given inputs, we identify the optimal sparse masks independently for each attention head. These head-specific masks are then combined as the sparse mask of entire multi-head attention. This finer granularity broadens the representational scope of sparse masks, thereby facilitating the capture of distinct sparse patterns hidden in shadowy sparsity. First, determining the optimal sparse mask for one head is relatively straightforward, because it is not required to account for other hands. Second, this approach can lead to further computational savings, since a score that is critical for one head might not be necessary for another.

\noindent\textbf{MLP Block.} Although overall activations are scattered, variations in activation frequency and values highlight the relative importance of each activation. Taking this detailed information into consideration, we selectively apply a filter to the activated neurons, effectively treating those of less importance as inactive. Given that the contributions of these neurons to the final outcome are minimal, their exclusion has a negligible impact. We implement this process in a block-wise manner, resulting in a structured sparse pattern that is well-suited to the characteristics of the hardware.

\section{Sequence-oriented Predictor}
\label{sec:predictor}
Although the precise sparse patterns can emerge naturally from computation outcomes, this does not contribute to computational savings. To truly reduce computation from sparsity, it is necessary to identify sparse patterns before the actual computation. Utilizing low-rank neural networks to accurately predict these sparse patterns has been proven feasible in LLM inference~\cite{deja-vu,powerinfer,llm-in-a-flash,dynamic-sparse-attention}. Given the similarity in freezing the majority of model parameters, we adopt a similar neural-network-based approach. However, there are new challenges introduced in fine-tuning. First, the sequenced inputs could easily lead to an excessively large predictor, compromising efficiency. Second, the updates of trainable parameters introduce bias to the predictor inputs, adversely affecting accuracy.




\subsection{Criterion \rom{1}: Efficiency}

We develop a two-stage design that guarantees the predictor's efficiency when handling sequence. In stage one, the predictor processes each token individually, which keeps the predictor's size constrained to the dimension of a single token. In stage two, we consolidate these individual predictions into an aggregated one that represents the final sparsity for the token sequence. Building on this sequence-oriented design, the detailed structure of predictor is specified as follows:

\begin{figure}
    \centering
    \includegraphics[width=0.50\textwidth]{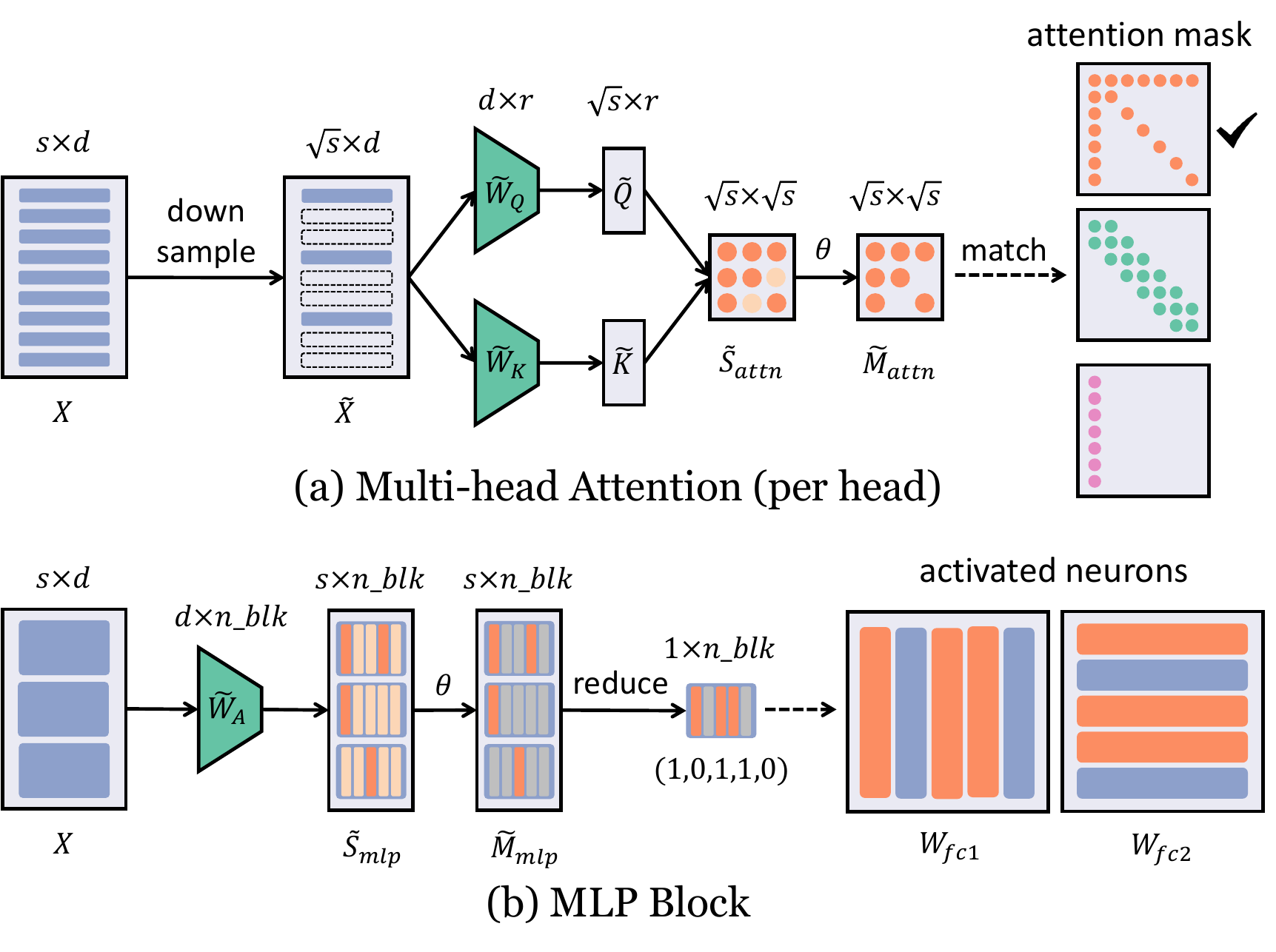}
    \caption{The process of prediction (batch size = 1 for simplicity). Each predictor comprises a set of trainable parameters (\textcolor{green}{green}), which accepts tensors (\textcolor{blue}{blue}) as input and produces approximations of attention scores or activations. These outcomes are then processed to filter out less significant values (\textcolor{lightorange}{light orange}), and a reduction is performed if needed, to generate the final sparse pattern (\textcolor{orange}{orange}) for the token sequence.}
    \label{fig:predictor}
\end{figure}

\noindent\textbf{Multi-head Attention.} Figure~\ref{fig:predictor}(a) shows the process of prediction in multi-head attention's one head. We construct a pair of trainable low-rank approximation matrices, $\hat{W_{Q}}$ and $\hat{W_{K}}$, to obtain the approximate queries $\hat{Q}$ and keys $\hat{K}$. Particularly, we down-sample the input $X$ in the sequence dimension, reducing its size from $s$ to $\sqrt{s}$ to decrease subsequent computation. This approach is rational because our primary concern lies with the overall distribution of attention scores for mask matching, rather than any individual value. The approximate attention scores are then calculated using the following formula:
\begin{equation*}
    \hat{S}_{attn}=\hat{Q}\hat{K}^{T}=X\hat{W_{Q}}X\hat{W_{K}}^T
\end{equation*}
Here $\hat{S}_{attn}\in\mathbb{R}^{\sqrt{s}\times\sqrt{s}}, \hat{W_{Q}}, \hat{W_{K}}\in \mathbb{R}^{d\times r}$ and $r \ll d$. When two approximation matrices are well trained, $\hat{S}_{attn}$ can provide a close estimation of accurate attention scores $S_{attn}$.

The $\hat{S}_{attn}$ is converted into a binary mask $\hat{M}_{attn}$ with a threshold. Additionally, a reduction in the batch dimension is performed to generate the sparse pattern of the whole input. The resulting binary mask is then categorized into one of several pre-defined typical masks. This strategy not only aligns the sparse pattern with expert insights~\cite{reformer,longformer,big-bird,informer,local-attn} but also provides convenience for the following efficient implementation.


\noindent\textbf{MLP Block.} Figure~\ref{fig:predictor}(b) shows our design of predictors for MLP block. Similarly, we construct a trainable low-rank matrix $\hat{W_{A}}$ to approximate the indices of the activated neuron blocks. Considering the correlated activation patterns of the two linear layers, the prediction result is applied to both layer weight matrices. The approximation can be expressed as:
\begin{equation*}
    \hat{S}_{mlp}=X\hat{W_{A}}
\end{equation*}
Here $\hat{S}_{mlp} \in \mathbb{R}^{s\times n\_blk}$ and $\hat{W_{A}}\in \mathbb{R}^{d\times n\_blk}$, where $n\_blk = \lceil d/blk\_size \rceil$ is the number of neuron blocks and $blk\_size$ is the number of neurons in each block. When $\hat{W_{A}}$ is well trained, $\hat{S}_{mlp}$ can indicate the importance of different neuron blocks in determining the final outputs for the given inputs.

The $\hat{S}_{mlp}$ is then binarized by applying a threshold, which serves to filter out blocks deemed less important. Finally, a reduction is performed on both the batch and sequence dimension of $\hat{M}_{mlp}$, to obtain the final sparsity pattern.

\subsection{Criterion \rom{2}: Accuracy}
All predictors are pre-trained offline using data collected from model inference. Since the size of predictors is relatively small, the training will come to convergence quickly and consume minimal resources compared to the following LLM fine-tuning. However, the updating of trainable parameters during fine-tuning could potentially skew the distribution of the original predictor inputs, thus impairing prediction accuracy. 

Consequently, we employ two optimizations during predictor training. First, we add noise to the original data for data argumentation, which helps the predictor avoid overfitting and boosts its robustness. Second, we prioritize recall over precision in the computation of prediction loss. This is because the final outcome is primarily affected when weights that should be active are incorrectly predicted as inactive.

\subsection{Analysis: Computational Savings and Overhead.}
We finally analyze the computational savings and associated overhead introduced by predictors. 

\noindent\textbf{Savings.} In the forward phase, the computational savings within multi-head attention are primarily derived from the computation of attention scores, which is reduced from $O(s^{2})$ to $O(s)$ thanks to the use of sparse masks. In MLP block, the computational savings depend on the sparsity ratio, scaling by an order of magnitude corresponding to $s$. In the backward phase, computational savings mirror those seen in the forward phase, as detailed in the analysis presented in Section~\ref{sec:opportunity}.

\noindent\textbf{Overhead.} In multi-head attention, the extra overhead for one batch primarily consists of three matrix multiplications:
\begin{equation*}
    \text{Cost}_{attn}=\text{Cost}_{Q}+\text{Cost}_{K}+\text{Cost}_{QK} = \sqrt{s}dr+\sqrt{s}dr+sr
\end{equation*}
In MLP block, the overhead for one batch arises from one matrix multiplication and one reduction operation:
\begin{equation*}
    \text{Cost}_{mlp}=\text{Cost}_{A}+\text{Cost}_{AND}=sdr+s
\end{equation*}
Given that $r \ll d$ and $d$ is a constant determined by model structure, the total computational complexity can be approximated as $O(s)$. Weighed against the substantial savings, the overhead introduced by predictor is considered acceptable.

\section{Dynamic-aware Operator}
\label{sec:operator}
Sparse linear algebra often struggles to match the performance of its dense counterparts due to irregular computation patterns and scattered memory accesses. Many sparsity tools~\cite{sparta,flash-llm,cusparse,sputnik} manage to achieve comparable performance based on fixed sparsity pattern, typically relying on either static kernel compilation or data format conversion before runtime. However, during fine-tuning, sparse patterns are highly input-dependent, which can only be determined at runtime. This dynamic nature of sparsity falls outside the capabilities of these sparsity tools. Fewer sparse tools~\cite{pit,megablocks} are tailored for dynamic scenarios. However, due to a lack of specific optimizations for our desired sparsity patterns, they only yield sub-optimal performance. To map the achieved sparsity onto hardware systems efficiently, we develop a suite of custom dynamic-aware operators. These operators make efficient use of predicted sparse patterns involved in the computation of both multi-head attention and MLP block.



\subsection{Multi-Head Attention}
\begin{figure}
    \centering
    \includegraphics[width=0.50\textwidth]{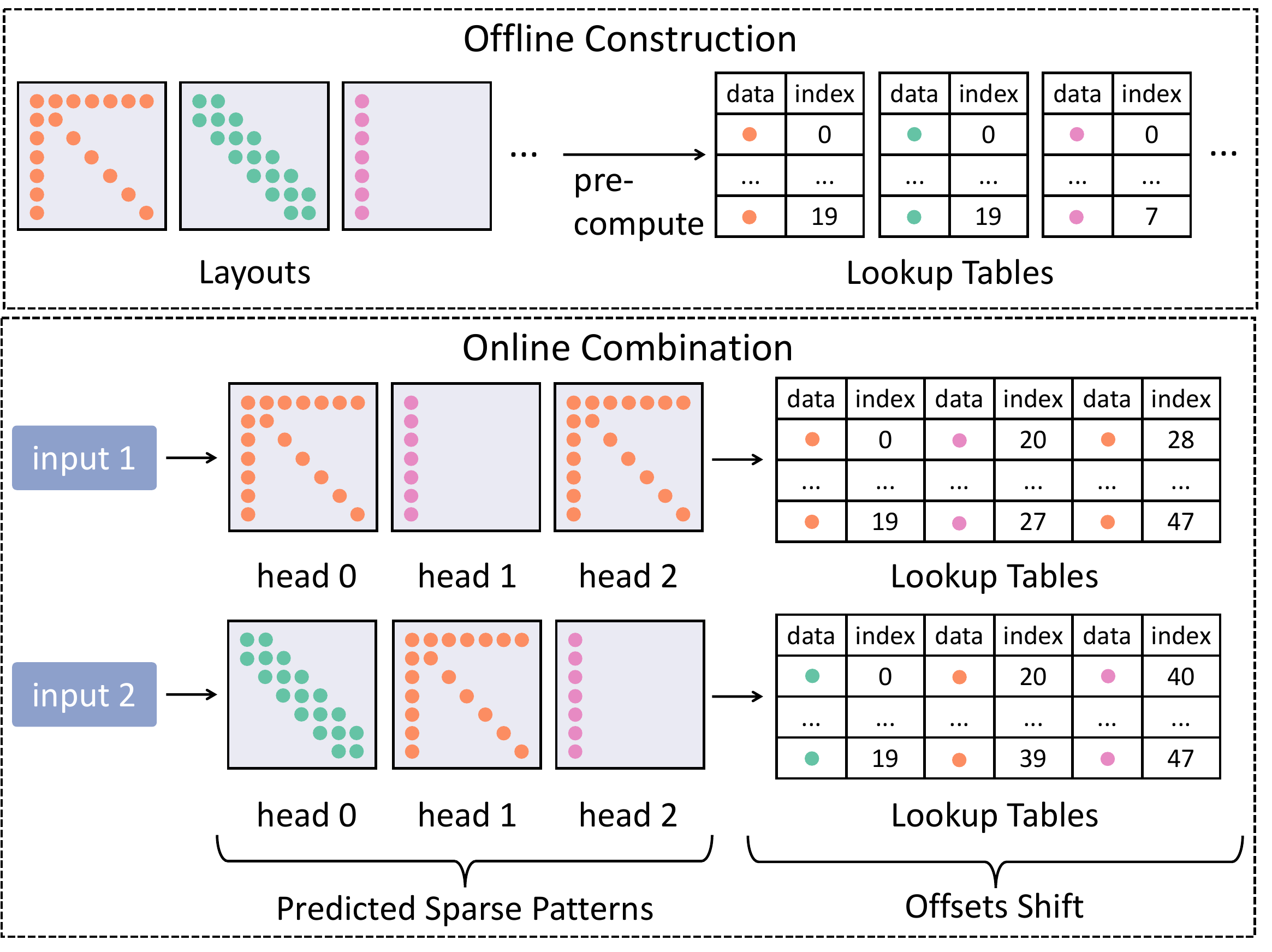}
    \caption{Example of two-stage approach in multi-head attention.}
    \label{fig:operator-attention}
\end{figure}

The computation of sparse attention can be broken down into two distinct block-wise sparse matrix multiplications: SDD and DSD, where `S' represents a sparse matrix and `D' denotes a dense matrix. A typical optimization is to pre-calculate the sparse pattern layout, which reduces the calculations required during runtime. However, the dynamic property of sparsity at runtime inherently clashes with the premise of pre-computation. To preserve computational efficiency in the face of dynamic sparsity, we propose a two-stage approach, performed offline and online, as shown in Figure~\ref{fig:operator-attention}.


\noindent\textbf{Offline Pool Construction.} The irregular nature of data layouts in sparse operations means that data indexing constitutes a significant computational workload. Pre-computing the data layout indices and storing them into lookup tables is crucial for achieving optimal performance in sparse operations. Rather than pre-computing certain fixed sparse pattern layouts—which is impractical due to the dynamic nature of the sparse patterns—we construct a pool of common atomic sparse patterns and pre-calculate their layouts. This strategy stems from the observation that existing sparse attention patterns often consist of a combination of these atomic patterns~\cite{attn-survey}.

\noindent\textbf{Online Pattern Combination.} Based on the predictor's output, each head is assigned a specific sparse pattern. These patterns will combined later to form the overall pattern of multi-head attention. During combination, only an offset needs to be added to the existing layout lookup tables. Finally, a list of data block indices is provided for sparse matrix multiplication. As the basic unit of operation is the block rather than the individual head, workload imbalance is avoided, even when the sparse patterns of different heads vary.

This two-stage approach enables a substantial portion of the computational workload to be shifted to the pre-runtime phase, while still preserving the flexibility needed for later integration to accommodate the demands of dynamic sparsity.


\subsection{MLP Block}
The sparsity introduced by ReLU within MLP block results in two sparse matrix multiplications occurring in both linear layers. Unlike common sparse operators, the sparse pattern here is uniquely column-wise or row-wise. It is because when an element within the MLP block's activation is zero, the corresponding column in the first linear layer as well as the row in the second linear layer are both rendered inactive. This particular characteristic allows us to craft two specific optimizations to achieve better performance.


\noindent\textbf{Neuron Sparsity.} Since the basic unit of activated weights is a column or a row, i.e. a neuron, we design a neuron-centric matrix multiplication based on the classic matrix multiplication tiling algorithm. Besides standard inputs, our specialized operator also accepts indices of activated neuron blocks. During computation, only the neuron blocks identified as active are loaded and computed. This approach is inherently compatible with the conventional tiling algorithm and eliminates the need for data format conversion.

\noindent\textbf{Memory Coalescing.} Another optimization arises from the data loading pattern of weights in the two linear layers: weights in the first layer are accessed column-wise, whereas those for the second linear layer are accessed row-wise. This inspires us to organize the weights in the two linear layers in a column-major format and a row-major format, respectively. This alignment with the memory access patterns of GPUs minimizes the overhead associated with data loading and enhances computational throughput.

\section{Evaluation}
\subsection{Experimental Setup}
\noindent\textbf{Machines.} We conduct experiments on two platforms, covering both data-center workstation and desktop professional GPU. Platform A contains an AMD EPYC 7V13 processor and an Nvidia A100 80GB GPU. The A100 GPU provides 1,555 GB/s memory bandwidth and 19.5 TFLOPs FP32 operations. Platform B contains an AMD EPYC 7742 processor and 4 Nvidia A6000 48GB GPUs. Each A6000 GPU offers 768 GB/s memory bandwidth and 38.71 TFLOPS FP32 operations.

\noindent\textbf{Models.} The models used for evaluation are detailed in Table~\ref{tab:model}. We choose models from two popular LLM families: OPT~\cite{opt} and GPT-2~\cite{gpt3}. For GPT-2, we mainly concentrate on the sparsity within multi-head attention, given that its activation function is GeLU~\cite{gelu}. Across all experiments, we employ mixed-precision techniques~\cite{mixed-precision} adhere to common practices, utilizing FP16 for parameters and FP32 for activations.

\begin{table}
    \small
    \centering
    \caption{Models for evaluation.}
    \label{tab:model}
    \begin{tabular}{llll}
    \hline
        Model & \# Params      & Batch Size & Seq Len  \\ \hline
          OPT & 350M/1.3B/2.7B & 2/4        & 512/1024 \\
        GPT-2 & 774M/1.5B      & 4/8        & 512/1024 \\
    \hline
    \end{tabular}
\end{table}

\noindent\textbf{PEFT Methods.} Our evaluation encompasses on three exemplary PEFT techniques: LoRA~\cite{lora}, adapter~\cite{adapter-1}, and bitfit~\cite{bitfit}, which are collectively the most widely utilized ones. Specially, we conduct our ablation studies and accuracy validation using LoRA owing to its broad acceptance within the domain.

\noindent\textbf{Datasets.} For performance evaluation, we apply a real-world dataset E2E~\cite{dataset-e2e} to ensure that the sparsity patterns within LLM reflect real-world situations. For accuracy validation, we first fine-tune the model on the widely used instruction dataset Alpaca~\cite{dataset-alpaca} and then evaluate its accuracy directly without further fine-tuning across a variety of representative downstream tasks, each providing unique challenges as detailed in Table~\ref{tab:downstream}.

\begin{table}
    \small
    \centering
    \caption{Downstream tasks for evaluation.}
    \label{tab:downstream}
    \begin{tabular}{ll}
    \hline
        Tasks & Description \\ \hline
        PIQA~\cite{dataset-piqa} & Physical commonsense reasoning \\
        Winogrande~\cite{dataset-winogrande} & Physical interactions understanding \\
        RTE~\cite{dataset-rte} & Natural language understanding\\
        COPA~\cite{dataset-copa} & Commonsense causal reasoning \\
        HellaSwag~\cite{dataset-HellaSwag} & Natural language commonsense \\
    \hline
    \end{tabular}
\end{table}

\noindent\textbf{Baselines.} We compare the overall performance of our system with the state-of-the-art fine-tuning library PEFT~\cite{baseline-peft}, which stands as the most relevant benchmark to our knowledge. Specially, to evaluate the performance of our design on multi-head attention, we draw comparisons with two classic sparse attention methods, Big Bird~\cite{big-bird} and Longformer~\cite{longformer}.

\subsection{Overall Performance}
\begin{figure}
    \centering
    \includegraphics[width=0.50\textwidth]{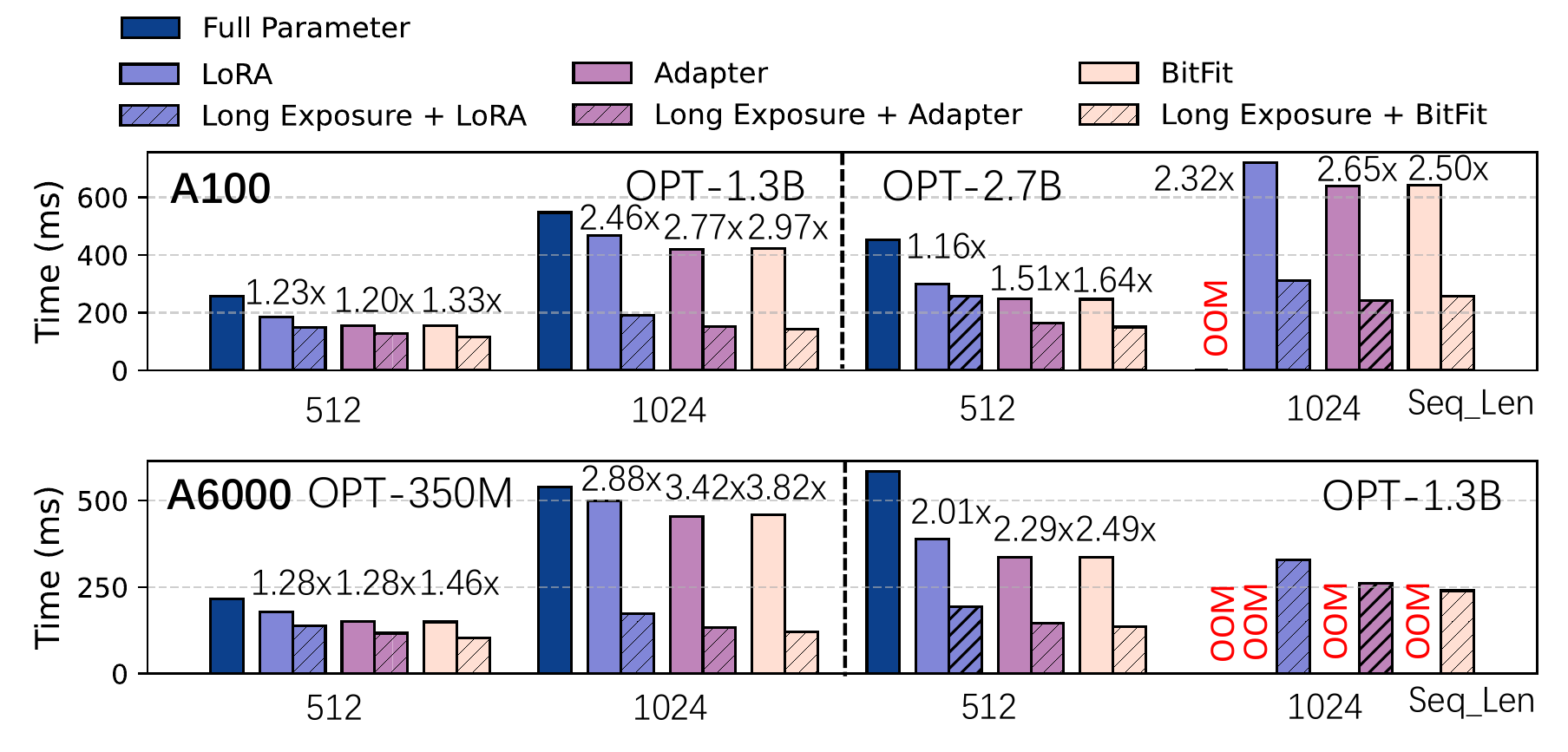}
    \caption{Execution time per batch and speedup of OPT.}
    \label{fig:exp-performance-time}
\end{figure}

\noindent\textbf{Execution Time.} We evaluate the execution time and corresponding speedup of \textsc{Long Exposure} on A100 and A6000 platforms, as shown in Figure~\ref{fig:exp-performance-time}. Integrating \textsc{Long Exposure} into three exemplary PEFT techniques, we examine two different parameter sizes and sequence lengths for each one. The results indicate that our system achieves up to $1.25\times$ speedup on average for OPT-1.3B with a sequence length of 512 on A100. As the sequence length doubles to 1024, the average speedup increases to $2.49\times$. This enhancement is attributed to the use of sparse attention masks, which alter the computation complexity from $O(s^{2})$ to $O(s)$. With a larger 2.7B model, the speedup remains consistent, averaging $1.44\times$ and $2.49\times$, respectively. Parallel results are observed on A6000, underscoring the robustness and reliability of our system. Since the introduction of sparsity reduces the execution time by decreasing the overall computation workload, this ensures a consistent speedup across different model sizes or platforms.

\begin{figure}
    \centering
    \includegraphics[width=0.50\textwidth]{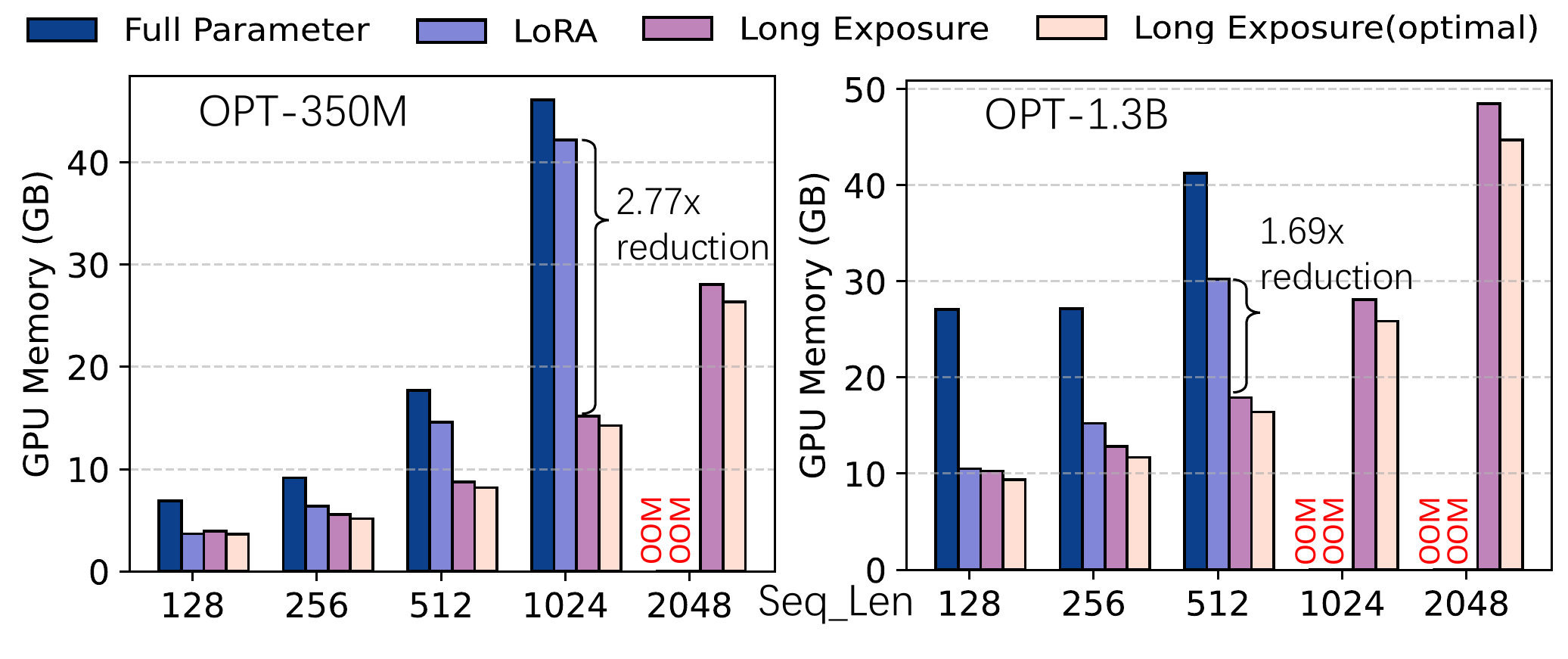}
    \caption{Memory footprints of OPT fine-tuning on A100.}
    \label{fig:exp-performance-memory}
\end{figure}

\noindent\textbf{Memory Footprint.} We also evaluate the memory footprints of \textsc{Long Exposure} as shown in Figure~\ref{fig:exp-performance-memory}. Despite not being explicitly designed for memory efficiency, the application of head-specific sparse attention masks alters the memory complexity from $O(s^{2})$ to $O(s)$, leading to lower memory footprints. Furthermore, selective activating model weights in MLP block permits the majority of the model to reside on the CPU, with only the active weights being transferred to the GPU for processing. This strategy can lead to additional memory savings, as presented by \textsc{Long Exposure} (optimal).

\begin{table}[b]
    \centering
    \small
    \caption{Comparative analysis of OPT model accuracy for downstream tasks after fine-tuning on the Alpaca dataset, with or without \textsc{Long Exposure}.}
    \label{tab:exp-performance-accuracy}
    \setlength{\tabcolsep}{2.0pt}
    \begin{adjustbox}{width=0.50\textwidth,center}
    \begin{tabular}{llrrrrrr}
    \hline
                                    &        & 350M-w/o & 350M-w  & 1.3B-w/o & 1.3B-w  & 2.7B-w/o & 2.7B-w  \\ \hline
        \multirow{2}{*}{PIQA}       & Acc.   & 65.13\%  & 64.80\% & 72.25\%  & 72.09\% & 74.70\%  & 73.45\% \\
                                    & Stderr & 1.11\%   & 1.12\%  & 1.05\%   & 1.06\%  & 1.02\%   & 1.02\%  \\
        \multirow{2}{*}{Winog.}     & Acc.   & 53.04\%  & 53.12\% & 58.88\%  & 58.80\% & 62.27\%  & 62.19\% \\
                                    & Stderr & 1.40\%   & 1.40\%  & 1.38\%   & 1.38\%  & 1.37\%   & 1.36\%  \\
        \multirow{2}{*}{RTE}        & Acc.   & 54.51\%  & 55.60\% & 54.15\%  & 54.51\% & 52.71\%  & 53.79\% \\
                                    & Stderr & 2.99\%   & 3.01\%  & 3.01\%   & 3.01\%  & 3.00\%   & 2.04\%  \\
        \multirow{2}{*}{COPA}       & Acc.   & 69.00\%  & 70.00\% & 81.00\%  & 81.00\% & 78.00\%  & 76.00\% \\
                                    & Stderr & 4.61\%   & 4.51\%  & 4.23\%   & 4.02\%  & 4.29\%   & 4.09\%  \\
        \multirow{2}{*}{Hella.}     & Acc.   & 32.26\%  & 32.40\% & 42.08\%  & 42.11\% & 46.76\%  & 43.95\% \\
                                    & Stderr & 0.47\%   & 0.47\%  & 0.499\%  & 0.49\%  & 0.50\%   & 0.50\%  \\ \hline 
    \end{tabular}
    \end{adjustbox}
\end{table}


\begin{figure*}
    \centering
    \includegraphics[width=0.98\textwidth]{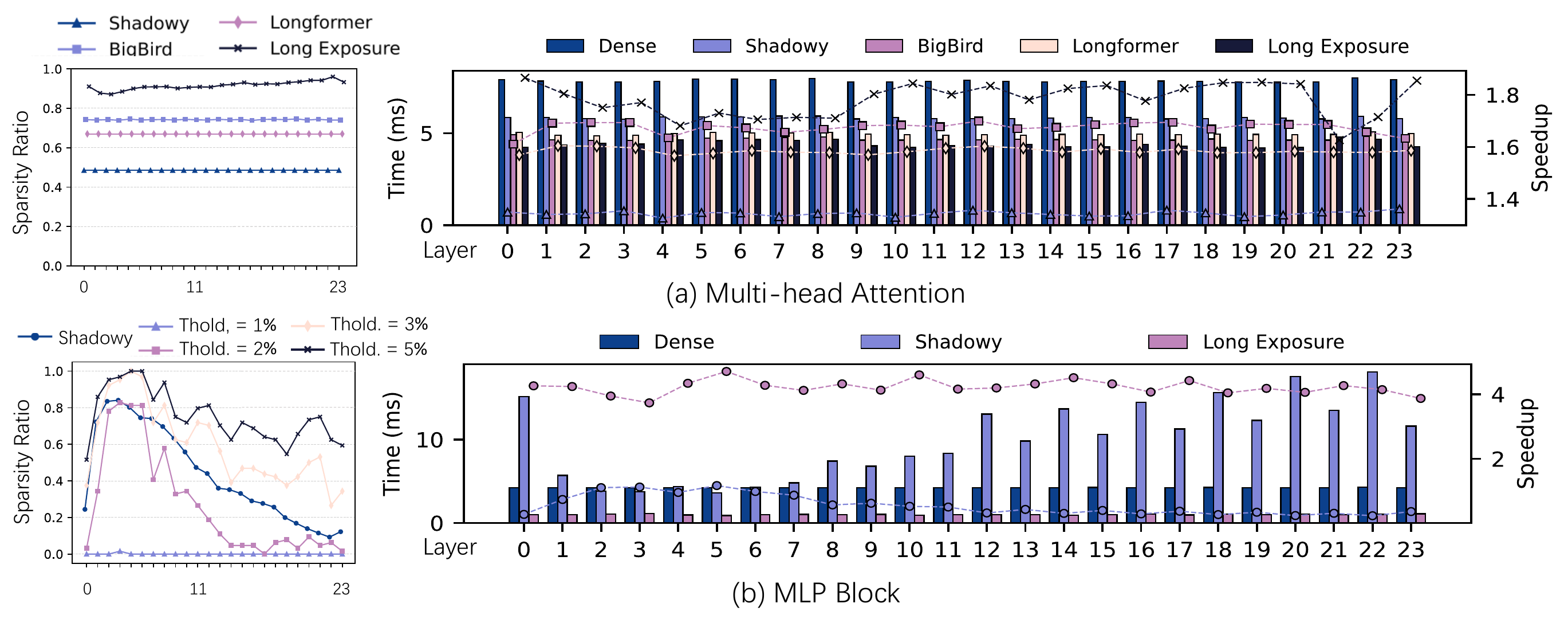}
    \caption{Sparsity ratio (left) and corresponding performance (right) of \textsc{Long Exposure} across different layers of OPT-1.3B. The term `Shadowy' denotes shadowy sparsity, which refers either to the sparsity in a uniform attention mask that covers all significant scores across all heads or to the sparsity present in the overall activations in MLP block.}
    \label{fig:exp-ablation-exposer}
\end{figure*}

\noindent\textbf{Model Accuracy.} We investigate the impact of \textsc{Long Exposure} on model accuracy by comparing with original LoRA across a variety of downstream tasks, as shown in Tabel~\ref{tab:exp-performance-accuracy}. We fine-tune OPT models of three distinct sizes on the Alpaca dataset. The results show that \textsc{Long Exposure} incurs only a minimal loss in downstream task accuracy across all model sizes and task types. This is because the essence of sparsity lies in disregarding the computation of elements that are zero or nearly zero, thereby only marginally affecting the final results.

\subsection{Ablation Study}
\begin{figure}
    \centering
    \includegraphics[width=0.50\textwidth]{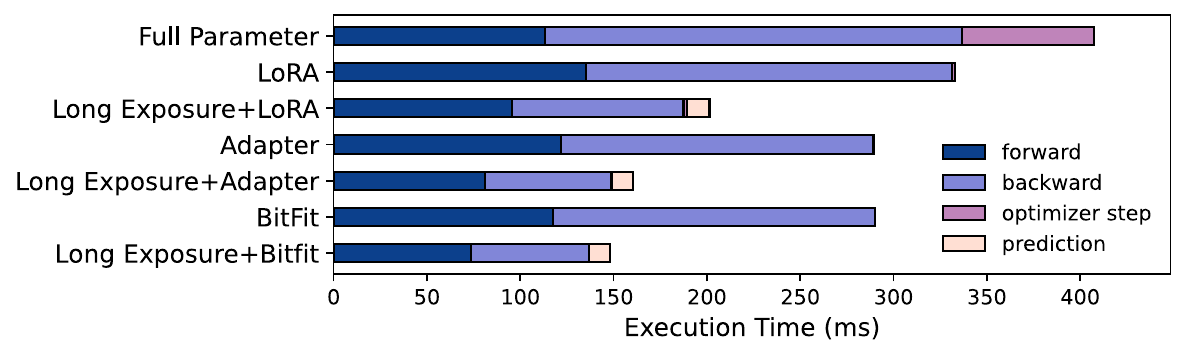}
    \caption{OPT-1.3B fine-tuning performance breakdown.}
    \label{fig:exp-ablation-breakdown}
\end{figure}

\noindent\textbf{Performance Breakdown.} Figure~\ref{fig:exp-ablation-breakdown} provides a detailed performance breakdown. We measure the execution time for three major phases within fine-tuning: the forward pass, the backward pass and the optimizer step. Additionally, we measure the prediction time in \textsc{Long Exposure} to evaluate the overhead introduced by predictors. The results show that compared to full fine-tuning, PEFT techniques can significantly reduce the execution time for optimizer step, while leaving the forward and backward passes largely unaffected. Building on this, \textsc{Long Exposure} achieves further reductions in execution time for both forward and backward passes across all three PEFT techniques. Although predictors are introduced to capture the sparsity patterns at runtime, their overheads are proved minimal, ensuring that the efficiency gains are preserved.

\noindent\textbf{Component \rom{1}: Shadowy-sparsity Exposer.} We begin by assessing the exposer's capability to capture model sparsity. Figure~\ref{fig:exp-ablation-exposer} (left) shows the sparsity ratios across different layers of OPT-1.3B when employing different methods. In multi-head attention, besides shadowy sparsity, we also use two traditional sparse attention methods, Longformer and Bigbird, as baselines. The results show that `shadowy' presents the lowest sparsity ratio. While Longformer and Bigbird can identify more sparsity, their use of a uniform attention mask results in a trade-off with accuracy. \textsc{Long Exposure} outperforms all other methods by employing more granular head-wise masks that are adept at revealing the sparsity concealed within shadowy sparsity. In MLP block, shadowy sparsity exhibits a relatively low sparsity ratio, typically not exceeding 60\% for most layers. \textsc{Long Exposure} utilizes a threshold-based filter to selectively ignore neurons that are activated but deemed less important. The results show that as the threshold (defined as a percentage of the peak values) is raised, the sparsity ratios increase correspondingly. By judiciously adjusting this threshold, \textsc{Long Exposure} strikes a balance between maintaining accuracy and enhancing efficiency.

In another view, we evaluate the corresponding performance gained from sparsity. Figure~\ref{fig:exp-ablation-exposer} (right) presents the execution time and speedups across different layers when fine-tuning OPT-1.3B. In multi-head attention, \textsc{Long Exposure} achieves $1.78\times$ speedup over the dense implementation and $1.33\times$ speedup over the `shadowy' method. In MLP block, \textsc{Long Exposure} outperforms the dense implementation with a speedup of $4.22\times$. Particularly, the 'shadowy' baseline exhibits lower performance compared to the dense implementation. This is attributed to its unstructured sparsity pattern, which differs from the structured block-wise sparsity utilized by \textsc{Long Exposure}, resulting in a reduced arithmetic intensity.

\begin{figure}
    \centering
    \includegraphics[width=0.50\textwidth]{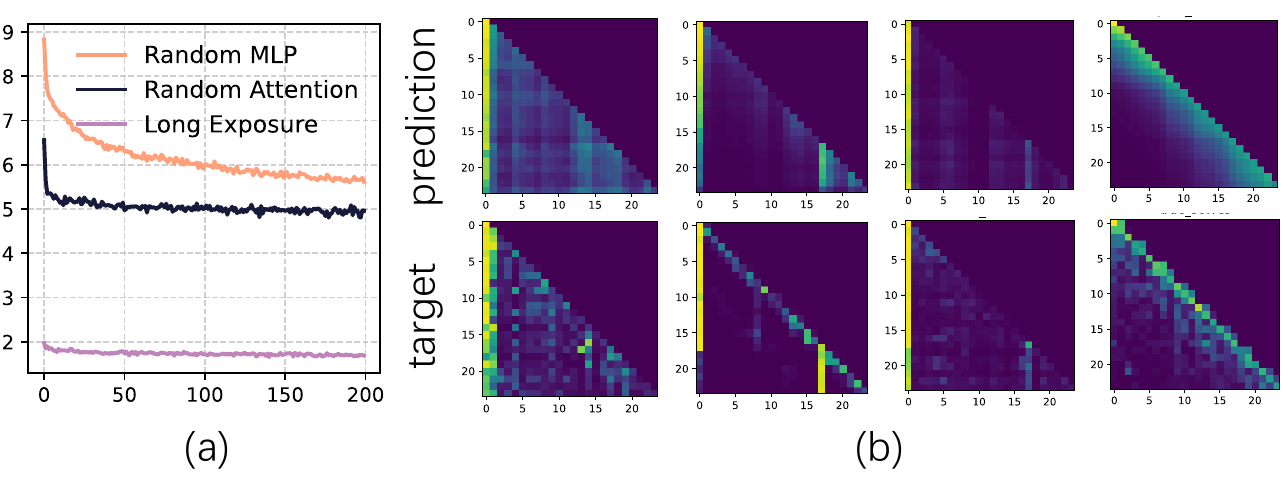}
    \caption{Fine-tuning loss curve (a) and prediction visualizations of predictors in multi-head attention (b).}
    \label{fig:exp-ablation-predictor}
\end{figure}

\noindent\textbf{Component \rom{2}: Sequence-oriented Predictor.}
We first evaluate the necessity of predictors. We compare the fine-tuning loss curves of our system with those of two baselines that employ random sparse patterns in multi-head attention and MLP block, respectively. As Figure~\ref{fig:exp-ablation-predictor}(a) shows, making accurate predictions of dynamic sparse patterns at runtime is crucial for model convergence with minimal loss. Beyond examining the loss curves, we also offer visual representations of the predictions from the multi-head attention's predictor. Figure~\ref{fig:exp-ablation-predictor}(b) shows that the predicted attention scores can closely approximate the ground truth for identifying the proper sparse pattern. Within MLP block's predictors, we report recall metrics, achieving an impressive average of $96.35\%$.

\begin{figure}
    \centering
    \includegraphics[width=0.50\textwidth]{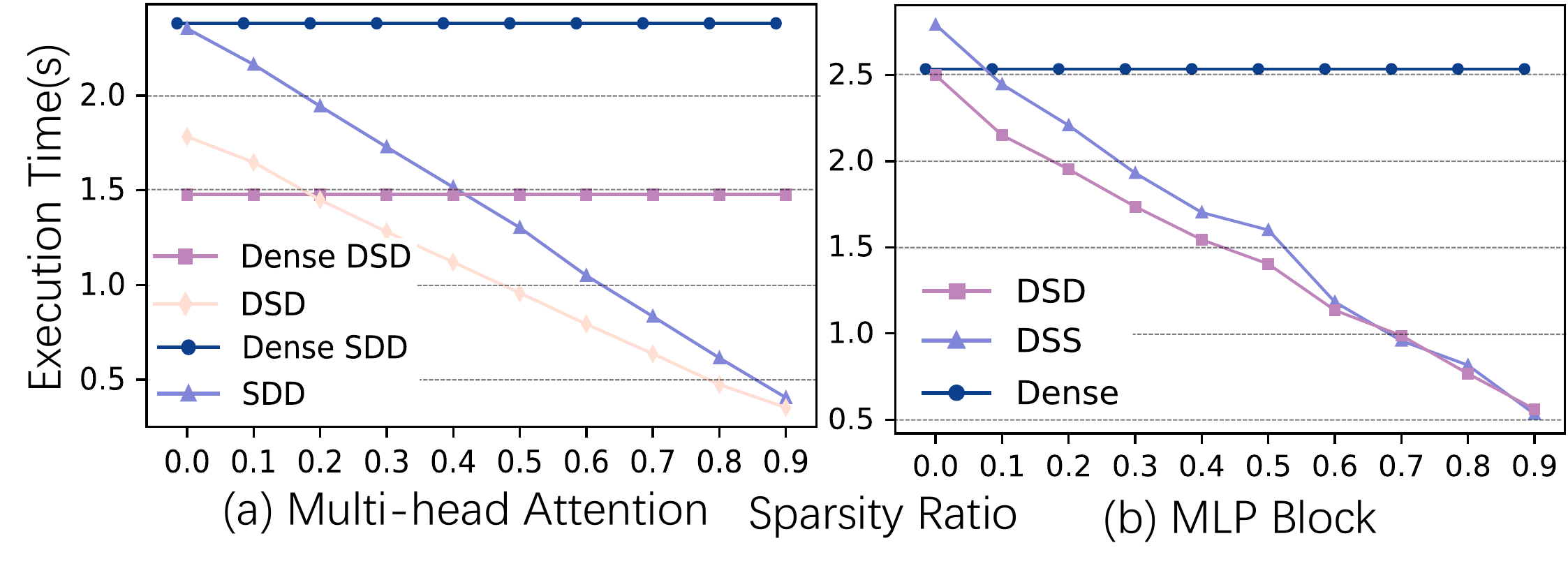}
    \caption{Dynamic operator performance compared to dense.}
    \label{fig:exp-ablation-operator}
\end{figure}

\noindent\textbf{Component \rom{3}: Dynamic-aware Operator.} We benchmark our operators against their dense counterparts under various sparsity ratios in Figure~\ref{fig:exp-ablation-operator}. In multi-head attention, the sparsity is applied block-wise, while in MLP block, the sparsity is neuron-wise. Both are aligned with the sparsity patterns utilized in \textsc{Long Exposure}. The results show that all operators can attain speedups as sparsity ratio increases, achieving enhancements of up to $3-5\times$. Besides, the execution time of all dynamic operators exhibits an almost linear relationship with sparsity ratio, suggesting that our operators are adaptable and efficient in scenarios with dynamic sparsity levels.

\subsection{Scalability}
\begin{figure}
    \centering
    \includegraphics[width=0.50\textwidth]{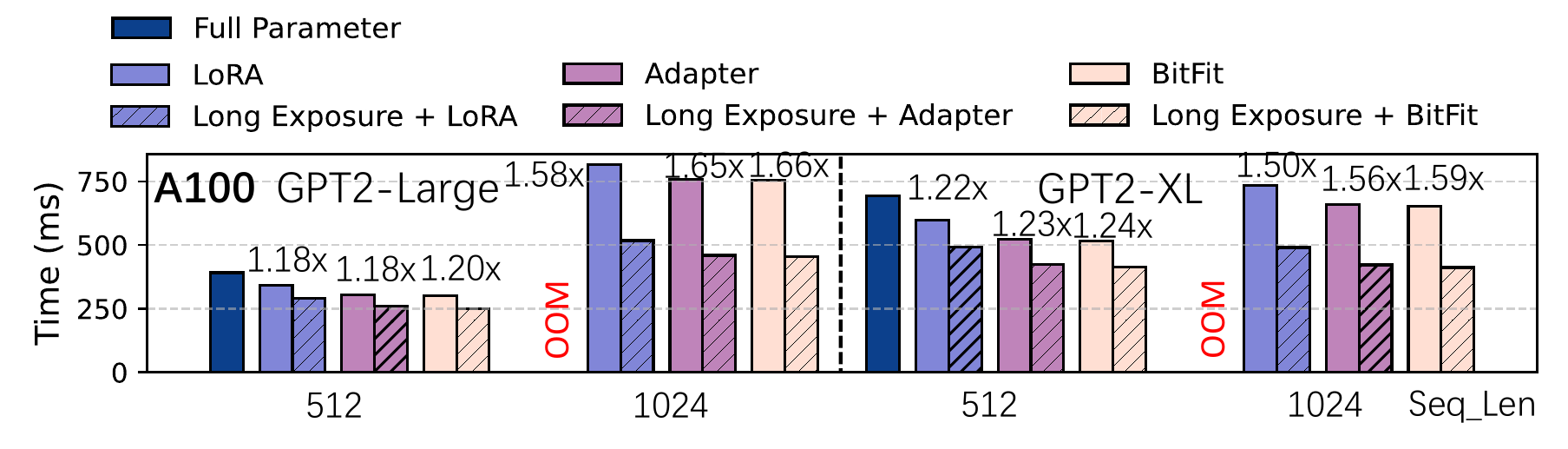}
    \caption{Execution time per batch and speedup of GPT-2.}
    \label{fig:exp-scale-card}
\end{figure}

\begin{figure}
    \centering
    \includegraphics[width=0.50\textwidth]{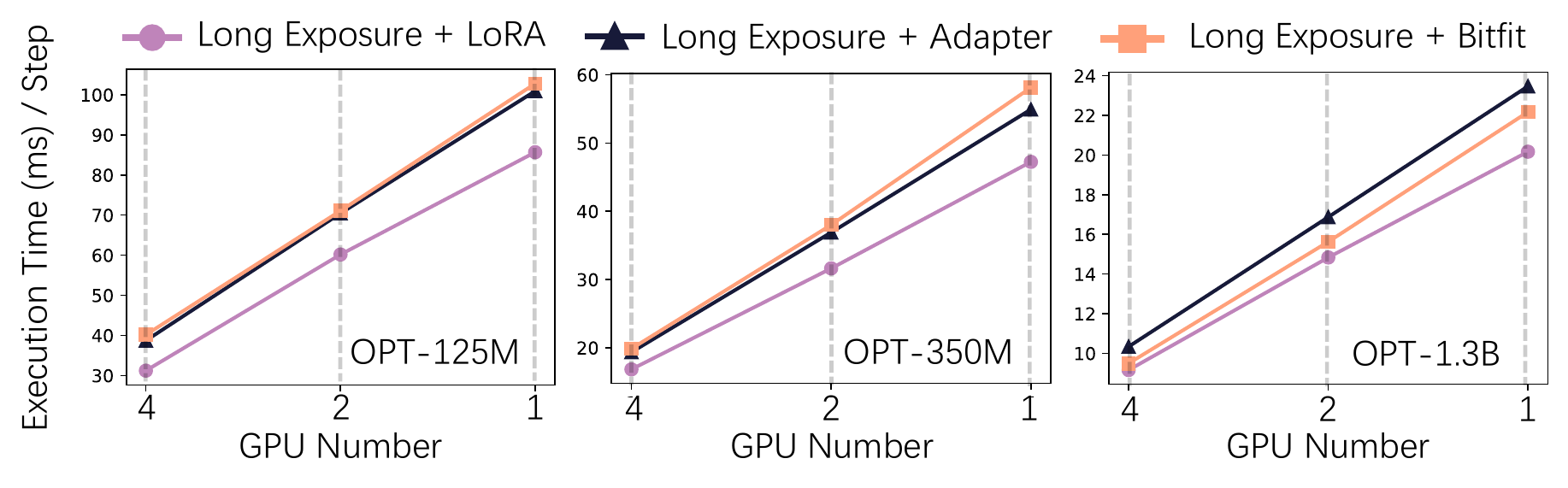}
    \caption{Strong scalability of \textsc{Long Exposure}.}
    \label{fig:exp-scale-model}
\end{figure}

We explore the scalability of \textsc{Long Exposure} from two aspects. The first is the model type. Beyond OPT, we extend our experiments to GPT-2, a GeLU-based model. As shown in Figure~\ref{fig:exp-scale-model}, although only optimizations on multi-head attention are applied, \textsc{Long Exposure} consistently achieves average speedups of up to $1.63\times$ and $1.55\times$ on two different model sizes, respectively. The second is the number of GPU utilized. We maintain a constant dataset size and increase the GPU count to evaluate the strong scalability of our system. Figure~\ref{fig:exp-scale-card} shows that the performance of our system scales linearly with the addition of more GPUs across three different model sizes. This is attributed to the fact that all optimizations within our system focus on the model computation workload, thereby introducing no extra communication overhead.

\section{Related Work}

\noindent\textbf{Model Pruning.}
Pruning aims to remove parameters without performance loss~\cite{pruning-survey}. Static pruning stands out as one such approach, which involves developing effective criteria for pruning models offline prior to subsequent inference~\cite{static-element, static-filter, static-group, static-sparse, static-deep-compression}. In contrast, dynamic pruning is conducted during runtime. Certain studies incorporate this process into the model training~\cite{dynamic-sft,dynamic-low-rank,dynamic-rl}, while others devise specialized optimizations for integration with fine-tuning~\cite{dynamic-movement,dynamic-rethink}. Different from these pruning techniques, \textsc{Long Exposure} does not prune any parameters but selectively activates a portion of them, preserving the model's original capacity for generalization. Furthermore, \textsc{Long Exposure} specifically aims to accelerate the fine-tuning process itself rather than the subsequent inference.

\noindent\textbf{PEFT Optimization.}
The widespread adoption of PEFT techniques has spurred numerous studies focused on optimization. Some efforts aim to enhance PEFT performance. AdaLoRA~\cite{adalora} proposes to adaptively allocate the parameter budget by significance. LoHa~\cite{loha} seeks to use more low-rank matrices for approximation and combines them with the Hadamard product. Some studies merge model compression with PEFT to improve further inference efficiency. Methods like QLoRA~\cite{qlora}, SPA~\cite{spa}, PST~\cite{pst}, LRP~\cite{lrp} combine quantization or model pruning with PEFT. In addition, there are studies dedicated to optimizing PEFT's memory footprint. LST~\cite{lst} introduces a ladder-side network for less gradient calculation. LoRA-FA~\cite{lora-fa} opts to freeze the down-projection matrix in LoRA for less activation memory. However, these studies primarily focus on algorithmic-level optimizations and overlook the wall-clock time impact of PEFT. \textsc{Long Exposure} accelerates PEFT holistically by addressing both algorithmic-level and system-level optimizations.

\section{Conclusion}
We propose \textsc{Long Exposure}, a highly efficient system designed to accelerate parameter-efficient fine-tuning for LLMs. Our approach notably identifies the intrinsic sparsity within LLM fine-tuning and introduces three key components that systematically capture, predict, and exploit these sparse patterns. \textsc{Long Exposure} demonstrates up to $2.49\times$ speedup over state-of-the-art methods, underscoring the potential for more extensive exploitation of sparsity for PEFT acceleration.

\section{Acknowledgment}
The authors express their gratitude to the anonymous reviewers for their insightful and constructive feedback. The work of Ju Ren was supported in part by the National Key R\&D Program of China under Grant No. 2022YFF0604502, the National Natural Science Foundation of China under Grant No. 62122095, 62341201 and 62072472, and by a grant from the Guoqiang Institute, Tsinghua University.

\bibliography{refer}
\bibliographystyle{IEEEtran}

\end{document}